\documentclass{article}
\usepackage{spconf,amsmath,epsfig}
\usepackage{graphicx}
\usepackage{pifont}
\usepackage{multirow}
\usepackage{subfigure}

\let\OLDthebibliography\thebibliography
\renewcommand\thebibliography[1]{
  \OLDthebibliography{#1}
  \setlength{\parskip}{0pt}
  \setlength{\itemsep}{0pt plus 0.3ex}
}

\begin{document}\sloppy

\def\x{{\mathbf x}}
\def\L{{\cal L}}

\title{Focus on Local Regions for Query-based Object Detection}
\name{Hongbin Xu, Yamei Xia, Shuai Zhao, Bo Cheng}
\address{
    School of Computer Science, Beijing University of Posts and Telecommunications\\
    \{xhb1806865437,ymxia,zhaoshuaiby,chengbo\}@bupt.edu.cn
}
\maketitle
\begin{abstract}
Query-based methods have garnered significant attention in object detection since the advent of DETR, the pioneering query-based detector. However, these methods face challenges like slow convergence and suboptimal performance. Notably, self-attention in object detection often hampers convergence due to its global focus. To address these issues, we propose FoLR, a transformer-like architecture with only decoders. We improve the self-attention by isolating connections between irrelevant objects that makes it focus on local regions but not global regions. We also design the adaptive sampling method to extract effective features based on queries' local regions from feature maps. Additionally, we employ a look-back strategy for decoders to retain previous information, followed by the Feature Mixer module to fuse features and queries. Experimental results demonstrate FoLR's state-of-the-art performance in query-based detectors, excelling in convergence speed and computational efficiency.
\end{abstract}

\section{Introduction}
\label{sec:intro}
Generic object detection aims at locating and classifying existing objects in any one image and labeling them with rectangular bounding boxes to show the confidences of existence.\cite{survey} The frameworks of these methods can mainly be categorized into two types: one-stage methods like SSD \cite{SSD} and YOLO \cite{YOLO}, and multi-stage detectors such as Faster R-CNN \cite{FasterRCNN}. Recently, transformer-based object detectors have gained attention,  with DETR\cite{DETR} introducing the transformer's encoder-decoder module to object detection. Unlike traditional detectors, DETR eliminates the need for anchor box design, relying on a set of learnable vectors called object queries for detection. However, DETR faces challenges in slow convergence and performance issues. Specifically, achieving results comparable to previous detectors on the COCO 2017 dataset requires over 500 training epochs for DETR, compared to the typical 12 epochs for Faster R-CNN.

To tackle this challenge, some studies \cite{DeformableDETR, AnchorDETR} have been proposed to address the problems. However, these approaches often bring in numerous extra parameters, enlarging the network and substantially raising training costs. Consequently, striking a suitable balance between convergence speed and computational complexity remains a challenge. Moreover, the attention mechanism can quickly establish correlations between unrelated objects, negatively impacting algorithm convergence. This has prompted our investigation and enhancement of the attention mechanism.
\begin{figure}[htbp]
    \centering
    \subfigure[origin figure]{\includegraphics[width=0.45\columnwidth]{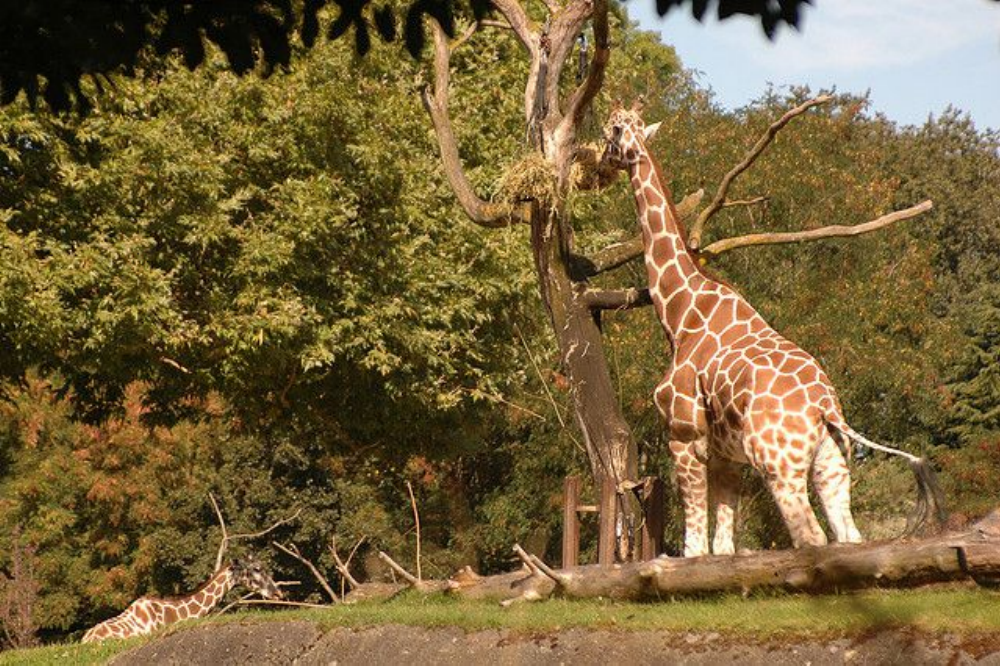}}\hspace{1pt}
    \subfigure[figure with attention]{\includegraphics[width=0.47\columnwidth]{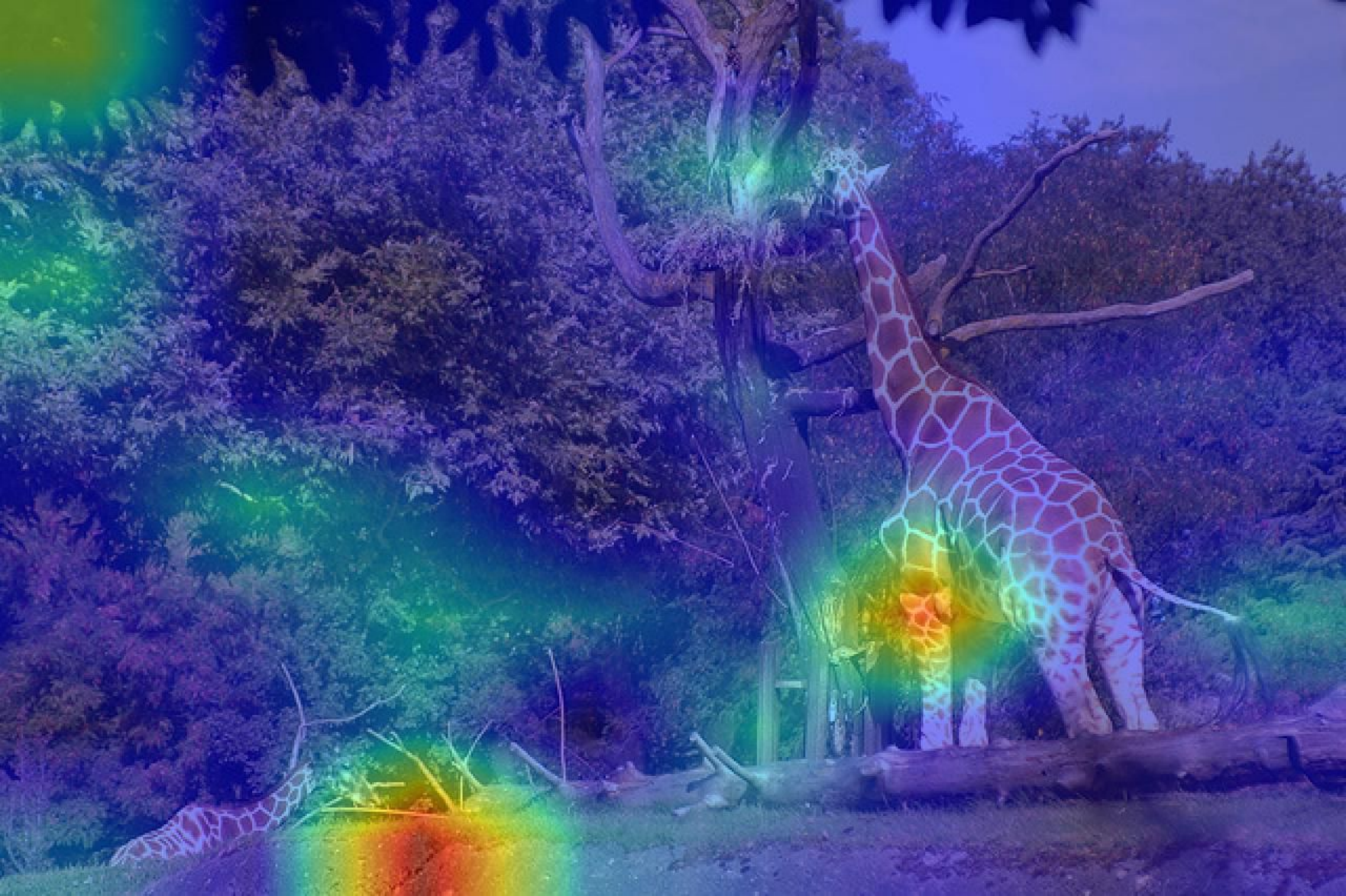}}
\caption{The outcome of visualizing attention indicates that, when pre-trained, pixels representing distinct entities may become associated despite having no inherent relationship.}
\label{fig:figure_attention_with_global}
\end{figure}

This paper introduces FoLR, an innovative query-based detector designed to address the mentioned challenges. In summary, our work contributes:
\begin{itemize}
\item Introducing FoLR, a novel query-based object detection method with a simplified decoder architecture that enhances self-attention with local regions.
\item Designing and implementing the Feature Mixer module and adaptive sampling method using DCN \cite{DCN} to boost feature-query interaction.
\item Highlighting FoLR's superiority with experimental results on the COCO dataset, it effectively overcomes the slow convergence and high computational cost limitations of traditional methods.
\end{itemize}
\section{Related Work}
\label{sec:related}
\subsection{Dense Methods}
In the early stages of deep learning-based object detection, anchor-based strategies were prevalent, involving the establishment and iterative refinement of anchor boxes through neural network training. Anchor-based approaches can be classified into one-stage (e.g., RetinaNet \cite{RetinaNet}, SSD \cite{SSD}) and multi-stage (e.g., Faster R-CNN \cite{FasterRCNN}) methods.

Dense methods, such as DenseBox \cite{DenseBox} and FCOS \cite{FCOS}, have emerged to address the challenge of accurately configuring anchor boxes, a task traditionally challenging for humans. These methods, collectively known as dense methods, leverage neural networks to autonomously learn object positions and shapes, generating numerous anchor boxes or candidate key points for each object.

\subsection{Query-Based Methods}
In contrast to dense methods, query-based detectors leverage the strengths of the transformer model and aim to decrease the number of candidates associated with each object. DETR \cite{DETR} is a pioneering end-to-end query-based algorithm that fully utilizes the transformer model in object detection. However, it has faced challenges such as slow convergence and suboptimal accuracy performance. To overcome these limitations, researchers have dedicated efforts to explore and propose various advancements in the field. For instance, Deformable DETR \cite{DeformableDETR} incorporates the idea of Deformable Convolutional Networks \cite{DCN} to optimize the attention module which significantly improves performance by enhancing the handling of object variations. Sparse R-CNN \cite{SparseRCNN} combines the query-based approach with Cascade R-CNN \cite{CascadeRCNN} while dispensing with the intricate encoder module of DETR. This combination results in accelerated convergence and improved accuracy. Building upon Sparse R-CNN, AdaMixer \cite{Adamixer} introduces an adaptive spatial 3D sampling method and the adaptive mixing procedure that enhances the interaction between queries and sampled features inspired by MLP-mixer \cite{MLPmixer}. 

In contrast to the methods mentioned above, our LoFR adopts a decoder architecture that improves the self-attention with local regions as well as utilizes the Feature Mixer module and the adaptive sampling method to promote the interaction and integration of features and queries.

\section{Method}
\label{sec:method}
\begin{figure*}[ht]
    \centering
    \includegraphics[width=\linewidth]{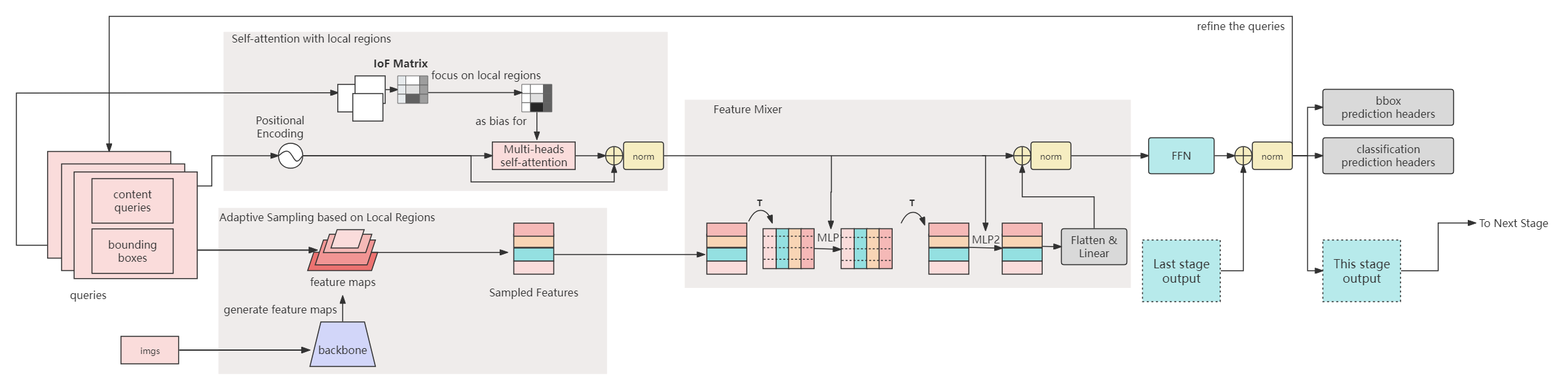}
    \caption{Overview of FoLR's decoder. The gray box represents the components of each stage decoder, comprising M stages. Similar to the Cascade Head \cite{CascadeRCNN}, each stage of the decoder refines the bounding boxes and content vectors obtained from the previous stage, progressively improving the accuracy of the prediction results.}
    \label{fig:FoLR_decoder_flow}
\end{figure*}
\subsection{Decouple Classification and Localization}
\label{sec:3.1}
In our proposed approach, we leverage the concept of the query that consolidates both semantic and positional information for each object, as seen in transformer-like algorithms like Sparse R-CNN \cite{SparseRCNN}. Previous studies \cite{DabDETR,Adamixer} demonstrate that decoupling the query into these two vectors significantly enhances model's recognition efficiency. We extend this approach by decomposing the query into two distinct vectors: the \textbf{content vector} and the \textbf{bounding box vector}. In the subsequent sections, we will use this naming convention to formally elucidate our approach.

\subsection{Self-Attention with Local Regions}
\label{sec:3.2}
Self-attention, frequently used in transformer-like methods to enhance query's descriptions, builds relationships within sequences. In object detection, it strengthens connections between queries, improving the descriptive ability of the content vector. However, our research finds that this global-focused attention mechanism creates unnecessary connections between unrelated queries (as shown in Fig. \ref{fig:figure_attention_with_global}). This hampers the detection task and negatively affects the model.

To enhance the model's ability to identify and filter out negative connections, we promote self-attention focusing on local regions. Initially, we compute the intersection of foreground (IoF) matrix between each query and others using Equation \ref{eq1}. If any matrix element falls below the specified threshold $\varepsilon$, we substitute it with a large negative number. This indicates that the information is unfavorable for self-attention and should be disregarded in subsequent operations.
\begin{equation}
\label{eq1}
\beta_{ij} =
    \left\{
        \begin{array}{lc}
            \log{\left(\frac{|box_i| \cap |box_j|}{|box_i|} + \sigma\right.)} & ,{if\; \beta_{ij} \geq \varepsilon}\\
            \\
            -infinity & ,{Otherwise}\\
        \end{array} 
    \right.
\end{equation}
where, $\beta \in \mathcal{R}^{(N_q, N_{q})}$ is indexed by $i$ and $j$, $N_q$ is the number of the queries, and $\sigma$ is a small constant set to $10^{-7}$, $\varepsilon$ is a constant taken from the interval $[0,1]$. 

\subsection{Extract Multi-Scale Features}
\label{sec:3.3}
To enhance the effectiveness of each query, it is crucial to extract meaningful and rich features which are generated by the backbone network from each input image. In this section, we introduce an adaptive sampling method inspired by Deformable Convolutional Networks \cite{DCN} for features extraction.

\noindent\textbf{Adaptive Sampling Method (ASM)}.
Given the $i$-th bounding box, we utilize the following equation to generate the $j$-th corresponding sampling points:
\begin{equation}
\label{eq2}
\left\{
     \begin{array}{l}
         \left\{\Delta x_{ij}, \Delta y_{ij}\right\} = Linear(Q)\\
         \widetilde{x_{ij}}=x_{ci}+\Delta x_{ij} \cdot w_i\\
         \widetilde{y_{ij}}=y_{ci}+\Delta y_{ij} \cdot h_i\\
     \end{array}
\right.
\end{equation}
where, $Q$ represents the content vector. $\Delta x_{ij}$ and $\Delta y_{ij}$ denote the offsets of the $j$-th sampling point relative to the center point $(x_{ci}, y_{ci})$ of the $i$-th bounding box, $w_i$ and $h_i$ represent the width and height of the $i$-th bounding box, respectively. 

\noindent\textbf{Focus on Local Regions}.
Now that we have acquired sampling points on the feature maps, the subsequent step for Sparse RCNN \cite{SparseRCNN} and AdaMixer \cite{Adamixer} involves interactively fusing these sampling points with queries. However, we contend that the features collected in this manner are derived from feature maps, not queries. Similar to DETR's decoder applying positional encoding on object queries, our method aims to improve the collected sampling points for better integration with queries. Here, we use a set of parameters generated by the linear layer to enhance the sampling points:
\begin{equation}
\label{eq3}
\left\{
     \begin{array}{l}
         \left\{\alpha^x_{ij}, \alpha^y_{ij}\right\} = Linear(Q)\\
         \widetilde{x_{ij}}=x_{ci}+\alpha^x_{ij}\\
         \widetilde{y_{ij}}=y_{ci}+\alpha^y_{ij}\\
     \end{array}
\right.
\end{equation}

In our approach, we employ the bilinear interpolation method to handle decimal coordinates of sampling points. Additionally, we adopt a strategy similar to the multi-head mechanism used in self-attention to enhance the diversity of the sampling points. Specifically, we divide the feature map's dimensions, denoted as d, into $N_{h}$ heads, where each head is assigned $d/N_{h}$ dimensions. This division ensures that each head captures distinct aspects of the features. For this study, we maintain $d=256$ and $N_{h}=4$.

Indeed, we generate $N_s$ sampling points at each level of feature maps each query, amounting to a total of $LN_s$ sampling points. However, the sampling points generated by different queries or the sampling points generated by the same query in different feature maps should not be of same significance. We assign the adaptive and learnable weights for the $i$-th query, the weighting method is as follows:
\begin{equation}
\label{eq4}
    S = w\cdot h
\end{equation}
\begin{equation}
\label{eq7}
    W = Softmax(Linear(S))
\end{equation}
Here, S represents the bounding box area corresponding to each query, calculated as the product of its width and height. $W_i$ with shape $\mathcal{R}^{(L, N_{h}\times N_s)}$ signifies the weight of the $i$-th query across the $L$ levels of feature maps.

\subsection{Feature Mixer}
\label{sec:3.4}
The interplay between sampled multi-scale features and the content vector is crucial in object detection. Inspired by MLP-mixer \cite{MLPmixer} and AdaMixer \cite{Adamixer}, we introduce Feature Mixer, a MLP module, to enhance this interaction. Following MLP-mixer's approach, we employ content queries to train two MLP networks for mixing the last two dimensions of the sampled features. The results are then converted into the shape of content queries and added to them as shown in  Fig. \ref{fig:FoLR_decoder_flow}.

In our design, we create a direct link between each query and sampled features, ensuring that each sampled feature interacts with its corresponding query. This process eliminates ineffective bins and generates the ultimate object feature. To maintain a lightweight design, we use two consecutive 1$\times$1 convolutions with the GELU activation function for the interaction. The parameters of these convolutions are generated from the corresponding content vector.

\subsection{Overview}
\begin{figure}[htbp]
    \centering
    \subfigure[origin image]{\includegraphics[width=0.32\columnwidth]{figures/origin.pdf}}\hspace{1pt}
    \subfigure[stage 1]{\includegraphics[width=0.32\columnwidth]{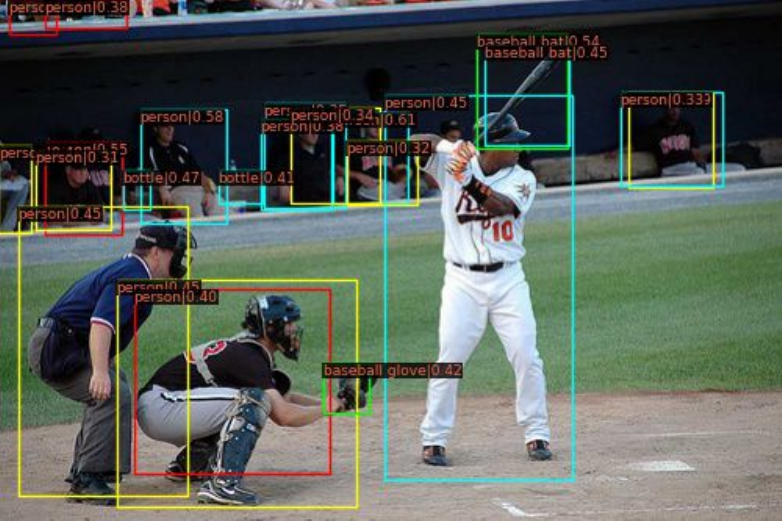}}\hspace{1pt}
    \subfigure[stage 6]{\includegraphics[width=0.32\columnwidth]{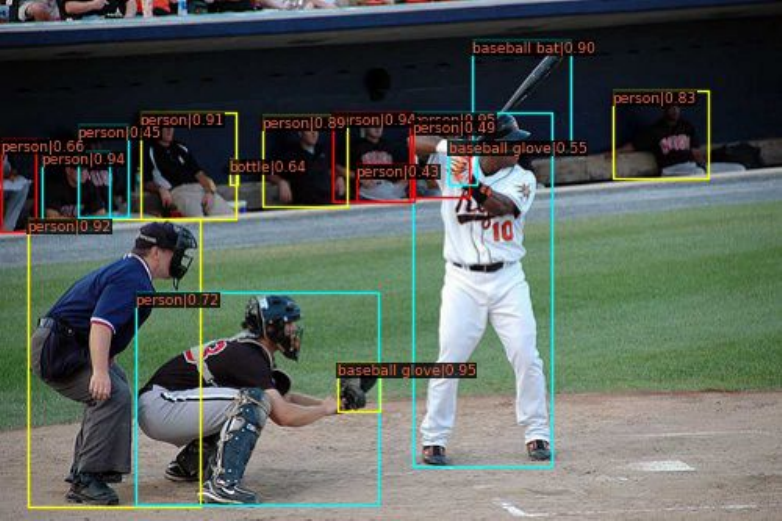}}
\caption{Detection results from various stages of FoLR with ResNet-50. Only the results of the three stages are shown here due to the space limitations.}
\label{fig:results_visualization}
\end{figure}

In this section, we provide a detailed introduction to our query-based detector, FoLR. FoLR is a fully decoder architecture comprising several key components: 
\begin{itemize}
\item Self-attention with local regions: Enhancing query descriptive capacity while mitigating negative relationships between irrelevant queries.
\item Adaptive Sampling Method(ASM): Adaptive feature sampling from multi-scale feature maps based on query positions.
\item Feature Mixer: Facilitating interaction between queries and ASM-sampled features for effective information fusion.
\item Prediction headers: Responsible for classification and regression tasks.
\end{itemize}
The interplay of these components shapes the holistic architecture of FoLR, empowering it with the capacity to proficiently extract features and detect objects. A visual representation of the architecture is depicted in Fig. \ref{fig:FoLR_decoder_flow}. Additionally, FoLR adopts a multi-stage refinement strategy, inspired by Cascade R-CNN \cite{CascadeRCNN}. The query vectors can be iteratively refined at each subsequent stage to enhance the accuracy of object detection, as exemplified in Fig. \ref{fig:results_visualization}.

\noindent\textbf{Look Back Strategy.} Later stages of the DETR-like detector utilize object queries from previous stages, but they are only employed during calculation, resulting in the loss of some effective information. To mitigate this loss, we apply the look-back strategy and connect object queries from the two stages before and after using a simple equation represented by Equation \ref{eq5}.
\begin{equation}
    \label{eq5}
    Q_{curr} = Q_{curr} + C\cdot Q_{prev}
\end{equation}

\noindent\textbf{Loss.} The model's loss, calculated using the DETR method, comprises focal loss ($\mathcal{L}_{cls}$) for predicted classifications, L1 loss ($\mathcal{L}_{L1}$), and generalized IoU loss ($\mathcal{L}_{giou}$) for predicted box coordinates. The overall loss is given by:
\begin{equation}
    \label{eq6}
    \mathcal{L} = \lambda_{cls}\cdot\mathcal{L}_{cls} + \lambda_{L1}\cdot\mathcal{L}_{L1} + \lambda_{giou}\cdot\mathcal{L}_{giou}
\end{equation}

\section{Experiments}
\label{sec:experiments}
\subsection{Dataset and Evaluation Metrics}
We conducted experiments on the COCO 2017 dataset \cite{COCO}, a popular benchmark for object detection and the mean average precision (AP) on the validation split is usually used as the major metric. FoLR is trained on the train set($\sim$118k images), and evaluated on the validation set($\sim$5k images) and the test-dev set($\sim$41k images). To offer a comprehensive assessment of FoLR's performance, attention is given to AP$_s$ scores for small-sized objects, AP$_m$ scores for medium-sized objects, AP$_l$ scores for large-sized objects, and GFLOPs as the metric for computational cost.

\subsection{Training Details}
The initial learning rate is set to $2.5\times 10^{-5}$. Following Sparse R-CNN \cite{SparseRCNN}, bounding boxes are initialized to cover the entire images. The content vector $Q$ with shape $\mathcal{R}^{(N_q,C)}$ is initialized using the initial weights from PyTorch's embedding module. In the adaptive sampling method (ASM), the linear layer weights for generating $\Delta x$ and $\Delta y$ start with zeros, and biases are uniformly initialized between -0.5 and 0.5. This ensures an even distribution of offset points within $[-0.5,0.5]$. During training on 2 GeForce RTX3090 GPUs, the AdamW optimizer is used, and the mini-batch is set to 8.

\subsection{Main Result}
We conducted experiments to compare FoLR with other methods. It's crucial to note that the results for all other methods are directly obtained from their respective original papers. To ensure a fair comparison, we evaluated FoLR's training using two different strategies. Firstly, with 100 learnable queries and standard random horizontal flipping augmentation, we trained FoLR using $1\times$ training schedule, enabling a comparison with mainstream dense detectors like Faster R-CNN. Conversely, for a comparison with other query-based detectors following a different training scheme, we allocated 300 learnable queries and incorporated additional random crop and multi-scale augmentation with $3\times$ training schedule.

\begin{table*}[ht]
    \centering
    \caption{Comparisons with other different object detectors on the MS COCO 2017 validation set.}
    \resizebox{0.95\textwidth}{0.22\linewidth}{
    \begin{tabular}{lcc|cccccc|c}
        \hline
        Methods (with backbone) & Feature & epochs & AP & AP$_{50}$ & AP$_{75}$ & AP$_s$ & AP$_m$ & AP$_l$ & GFLOPs\\
        \hline
        Faster R-CNN-R50\cite{FasterRCNN}   & FPN & 12 & 37.4 & 58.1 & 40.4 & 21.2 & 41.0 & 48.1 & 207\\
        Cascade R-CNN-R50\cite{CascadeRCNN} & FPN & 12 & 40.3 & 58.6 & 44.0 & 22.5 & 43.8 & 52.9 & 235\\
        \textbf{FoLR-R50(ours)}             & ASM & 12 & \textbf{42.6} & \textbf{61.8} & \textbf{46.0} & \textbf{25.2} & \textbf{45.6} & \textbf{58.8} & \textbf{104}\\
        \hline
        Faster R-CNN-R101\cite{FasterRCNN}   & FPN & 12 & 39.4 & 60.1 & 43.1 & 22.4 & 43.7 & 51.1 & 287\\
        Cascade R-CNN-R101\cite{CascadeRCNN} & FPN & 12 & 42.0 & 60.4 & 45.7 & 23.4 & 45.8 & 55.7 & 315\\
        \textbf{FoLR-R101(ours)}             & ASM & 12 & \textbf{43.5} & \textbf{62.9} & \textbf{46.6} & \textbf{25.4} & \textbf{46.6} & \textbf{60.3} & \textbf{184}\\
        \hline
        Deformable DETR-R$50^{\dagger}$\cite{DeformableDETR} & DeformEncoder & 50 & 44.3 & 63.2 & 48.6 & 26.8 & 47.7 & 58.8 & 173\\
        Sparse RCNN-R$50^{\dagger}$\cite{SparseRCNN}         & FPN     & \textbf{36} & 45.0 & 64.1 & 48.9 & 28.0 & 47.6 & 59.5 & 174\\
        Anchor DETR-R$50^{\dagger}$\cite{AnchorDETR}         & DecoupEncoder & 50 & 42.1 & 63.1 & 44.9 & 22.3 & 46.2 & 60.0 & \textbf{116}\\
        DAB-DETR-DC5-R$50^{\dagger}$\cite{DabDETR}           & Encoder       & 50 & 44.5 & 65.1 & 47.7 & 25.3 & 48.2 & 62.3 & 194\\
        DN-DETR-R$50^{\dagger}$\cite{DNDETR}                 & Encoder       & 50 & 44.1 & 64.4 & 46.7 & 22.9 & 48.0 & 63.4 & 121\\
        AdaMixer-R$50^{\dagger}$\cite{Adamixer}              & \ding{56}     & \textbf{36} & \textbf{47.0} & 66.0 & 51.1 & \textbf{30.1} & 50.2 & 61.8 & 132\\
        \textbf{FoLR-R$\mathbf{50^{\dagger}}$(ours)}         & ASM    & \textbf{36} & 46.7 & \textbf{66.5} & \textbf{51.3} & 28.8 & \textbf{50.3} & \textbf{62.8} & 119\\
        \hline
        Sparse RCNN-R$101^{\dagger}$\cite{SparseRCNN}               & FPN     & \textbf{36} & 46.4 & 64.6 & 49.5 & 28.3 & 48.3 & 61.6 & 254\\
        Anchor DETR-R$101^{\dagger}$\cite{AnchorDETR}               & DecoupEncoder   & 50 & 43.5 & 64.3 & 46.6 & 23.2 & 47.7 & 61.4 & \textbf{196}\\
        DAB-DETR-DC5-R$101^{\dagger}$\cite{DabDETR}                 & Encoder         & 50 & 45.8 & 65.9 & 49.3 & 27.0 & 49.8 & 63.8 & 263\\
        DN-DETR-R$101^{\dagger}$\cite{DNDETR}                       & Encoder         & 50 & 45.2 & 65.5 & 48.3 & 24.1 & 49.1 & 65.1 & 200\\
        AdaMixer-R$101^{\dagger}$\cite{Adamixer}                    & \ding{56}       & 50 & \textbf{48.0} & 67.0 & \textbf{52.4} & \textbf{30.0} & 51.2 & 63.7 & 208\\
        \textbf{FoLR-R$\mathbf{101^{\dagger}}$(ours)}               & ASM    & \textbf{36} & 47.9 & \textbf{67.6} & 51.5 & 29.4 & \textbf{51.8} & \textbf{65.7} & 198\\
        \hline
    \end{tabular}}
    \label{table:Comparisons_with_other_detectors}
\end{table*}

\begin{figure}[htbp]
\centering
    \includegraphics[width=0.8\linewidth]{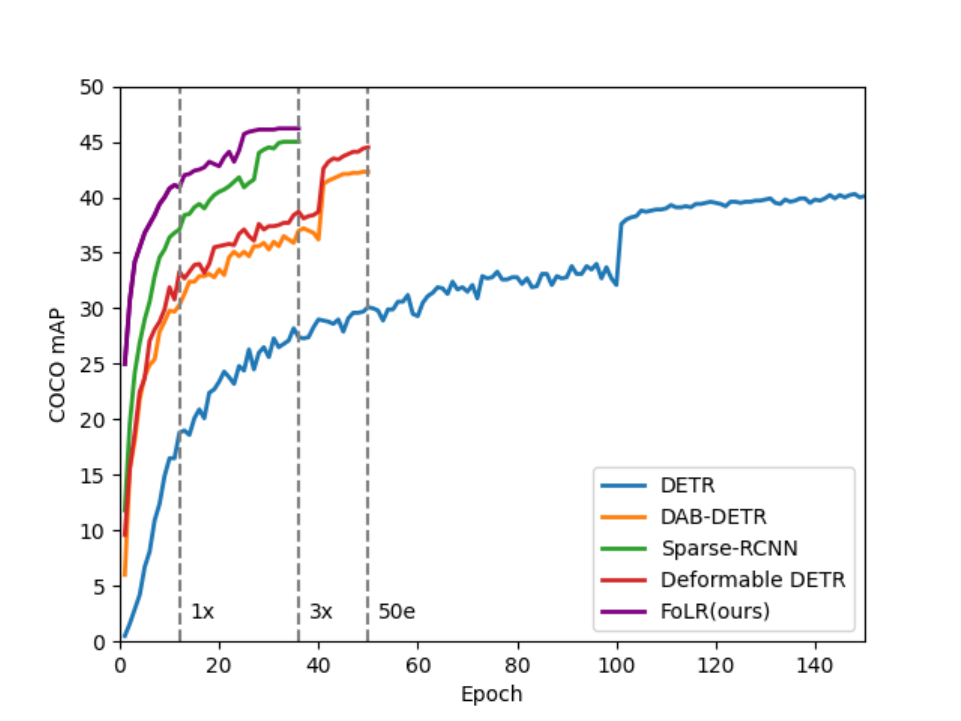}
    \caption{Convergence curves of FoLR and other detectors (ResNet-50 backbone) on the MS COCO 2017 val.}
    \label{fig:convergence_speed_curve}
\end{figure}
The final results are presented in Table \ref{table:Comparisons_with_other_detectors}. Remarkably, FoLR outperforms dense detectors like Cascade R-CNN in terms of Average Precision (AP), achieving 42.6 AP (vs. 40.3 AP) with ResNet-50 and 43.5 AP (vs. 42.0 AP) with ResNet-101 backbones. Additionally, FoLR attains the relatively high scores among query-based detectors, surpassing with 46.7 AP and 47.9 AP using ResNet-50 and ResNet-101 backbones, respectively. Furthermore, FoLR excels across other metrics, including accuracy for small object and computational cost. These results provide compelling evidence that FoLR effectively strikes a balance between complexity and performance.

Table \ref{table:Comparisons_with_other_detectors_test_dev} presents a comparison between FoLR and other methods on the COCO 2017 test-dev dataset. When employing ResNet-50, ResNet-101 as backbones, FoLR achieves AP scores of 47.1 and 48.0, respectively. These remarkable results reinforce the notion that FoLR's exceptional performance can also be extended and adapted to other datasets.

Moreover, as illustrated in Fig. \ref{fig:convergence_speed_curve}, we present a comprehensive analysis of FoLR's convergence speed by comparing it with other query-based detectors. The results clearly demonstrate that FoLR consistently outperforms the query-based detectors in terms of convergence speed at various stages of training. Specifically, FoLR demonstrates superior accuracy, achieving 46.7 AP (vs. Deformable DETR's 44.5 AP) while utilizing shorter training schedule of 36 epochs (vs. Deformable DETR's 50 epochs).
\begin{table}[htbp]
    \centering
    \caption{Comparisons with different methods on COCO test-dev set. The top section shows results from original papers.}
    \resizebox{\linewidth}{0.18\linewidth}{
    \begin{tabular}{lc|cc}
        \hline
        Method & Backbone & AP & AP$_s$\\
        \hline
        Deformable DETR\cite{DeformableDETR} & ResNet-50      & 46.9 & 27.7\\
        Sparse RCNN\cite{SparseRCNN}         & ResNeXt-101    & 46.9 & 28.6\\
        AdaMixer\cite{Adamixer}              & ResNet-50      & 47.2 & 28.3\\
        \hline
        \textbf{FuMA(ours)} & ResNet-50       & 47.1 & 28.9\\
        \textbf{FuMA(ours)} & ResNet-101      & 48.0 & 29.6\\
        \hline
    \end{tabular}}
    \label{table:Comparisons_with_other_detectors_test_dev}
\end{table}
\subsection{Ablation Studies}
Here, we also perform experiments to evaluate the effectiveness of the modules in FoLR. Due to computational constraints, we employ the ResNet-50 backbone and a $1\times$ training schedule for these experiments.

\noindent\textbf{Design of Attention with Local Regions}. We conduct additional experiments to validate the effectiveness of the attention module in the FoLR method. Different $\varepsilon$ values indicate varying degrees of attention to local regions. Specifically, when $\varepsilon$=0, it means the local regions strategy is not applied. Simultaneously, we assign distinct $\varepsilon$ values to the decoder across different stages. Setting $\varepsilon$ to 0.1, 0.2, and 0.4 in the last three stages, with the remaining stages at 0, leads to a performance improvement of up to 0.5. The results are detailed in Table \ref{table:strategies_in_Attention}. 
The experiments in Table 4 show 0.4 AP improvement with the Look-Back strategy compared to its absence.
\begin{table}
    \centering
        \begin{minipage}{0.49\columnwidth}
        \caption{Ablation Study of different $\varepsilon$ in the Attention.}
        \label{table:strategies_in_Attention}
        \centering
            \begin{tabular}{c|cc}
            \hline
                Methods & $\varepsilon$ & AP\\
            \hline
                \multirow{3}*{single} & 0 & 42.1\\
                ~ & 0.1 & 42.3\\
                ~ & 0.2 & 42.0\\
            \hline
                multiple &  & \textbf{42.6}\\
            \hline
            \end{tabular}
	    \end{minipage}
        \hfill
	\begin{minipage}{0.49\columnwidth}
		\centering
        \caption{Ablation Study of different strategies in FoLR.}
        \begin{tabular}{c|cc}
        \hline
            strategy & AP & AP$_s$\\
        \hline
            \ding{56}       & 41.9 & 24.9\\
            local regions & 42.4 & 24.8\\
            look back & 42.3 & 25.0\\
            both & \textbf{42.6} & \textbf{25.2}\\
        \hline
        \end{tabular}
        \label{table:strategies_in_FoLR}
	\end{minipage}
\end{table}
\begin{table}[htbp]
  \centering
  \caption{Ablation Study of different numbers of sampling points $N_s$ used in the ASM.}
  \resizebox{\linewidth}{0.13\linewidth}{
    \begin{tabular}{c|ccccccc}
        \hline
        N$_s$ & AP & AP$_{50}$ & AP$_{75}$ & AP$_s$ & AP$_m$ & AP$_l$ & Param\\ 
        \hline
        24 & 40.9 & 58.6 & 43.5 & 22.9 & 42.5 & 55.9 & 76M\\
        32 & \textbf{42.6} & \textbf{61.8} & \textbf{46.0} & \textbf{25.2} & \textbf{45.6} & \textbf{58.8} & 99M\\
        40 & 41.5 & 60.5 & 44.6 & 23.7 & 43.8 & 57.4 & 127M\\
        48 & 41.8 & 60.9 & 44.8 & 23.8 & 44.0 & 57.7 & 159M\\
        \hline
    \end{tabular}}
    \label{table:Ns}
\end{table}

\noindent\textbf{Design of Adaptive Sampling Method}. We also compare results obtained with a focus on local regions strategy against results without employing this strategy. The findings shown in Table \ref{table:strategies_in_FoLR} indicate that the focus on local regions strategy led to a performance improvement of 0.7 AP. Simultaneously, the choice of $N_s$ impacts both the convergence and the computational resources required by FoLR. To strike a balance, we conducted experiments with different values of $N_s$ in the range of $[24,48]$ with a step size of 8. The results in Table \ref{table:Ns} demonstrate that $N_s$=32 can achieve a good balance between performance and computational load.

\section{Conclusion}
\label{sec:conclusion}
In this paper, we introduce FoLR, a query-based decoder architecture that simplifies the transformer architecture, achieving faster convergence with reduced computational resource consumption. Notably, we improve the self-attention with global regions common in transformer decoders and propose an adaptive sampling method to gather effective features in feature maps, emphasizing local information based on queries. The Feature Mixer module facilitates efficient interactions between queries and feature maps. Importantly, a look-back strategy for decoders prevents forgetting of prior information. Through these innovations, FoLR outperforms recent detectors in both performance and computational resource efficiency.

\bibliographystyle{IEEEbib}

\end{document}